\documentclass{article}
\usepackage{iclr2027_conference,times}

\newif\ifarxiv
\arxivtrue
\ifarxiv\iclrfinalcopy\fi
\usepackage{amsmath,amssymb,amsthm}
\usepackage{graphicx}
\usepackage{booktabs}
\usepackage{microtype}
\usepackage[table]{xcolor}
\usepackage{array}
\usepackage{tabularx,longtable}
\usepackage{placeins}
\usepackage{flafter}
\usepackage{wrapfig}
\usepackage{needspace}
\usepackage{hyperref}
\usepackage{url}
\usepackage{tikz}
\usetikzlibrary{arrows.meta,positioning}

\newcommand{\method}{\mbox{HM-Router}}
\newcommand{\bench}{\textsc{hm-router-bench}}
\definecolor{oursblue}{HTML}{2878B5}
\definecolor{ourstint}{HTML}{E8F1FA}
\definecolor{tablegray}{HTML}{EFEFEF}
\newcommand{\best}[1]{\textbf{#1}}
\newcommand{\second}[1]{\underline{#1}}
\newcommand{\ours}{\textcolor{oursblue}{\method{}}}
\hypersetup{colorlinks=true,allcolors=oursblue}
\newcommand{\um}{\mathbf{u}_m}
\newcommand{\uh}{\mathbf{u}_h}
\newcommand{\qt}{\tilde{\mathbf{q}}}

\title{\method: Joint Model and Harness\\ Routing for Agentic Systems}
\author{Hao Mark Chen$^{1}$\quad Royson Lee$^{2}$\quad Yasuyuki Okoshi$^{3}$\quad Dimitris Anastasiou$^{1}$\\
\textbf{Wayne Luk$^{1}$\quad Hongxiang Fan$^{1}$}\\[0.6ex]
\normalfont $^{1}$Imperial College London\quad $^{2}$Samsung\quad $^{3}$Institute of Science Tokyo}

\begin{document}
\maketitle
% iclrfinalcopy would print "Published as a conference paper at ICLR 2027".
\ifarxiv\lhead{Preprint}\fi

\begin{abstract}
Agent performance depends on both the underlying model and the harness that manages its tool use and execution. Selecting a suitable pair requires accounting for their compatibility, yet training samples may cover only a subset of the growing combination space. We introduce \method{}, a routing method that jointly selects a model and harness for each query. It learns separate model and harness representations shared across routes, with an interaction term inspired by canonical polyadic (CP) tensor decomposition to capture how their compatibility varies with the query. This sharing allows training samples from observed pairs to inform predictions for unobserved combinations. We curate \bench{} from 12 public agent benchmarks, covering 293 routes, 73 models, and 25 harnesses. \method{} exceeds the strongest evaluated learned baseline by 7.3 percentage points in mean routing accuracy and leads at all seven evaluated cost budgets on the six-benchmark \textsc{HM-Route-cost} subset. When 90\% of routes have their training outcomes withheld, allowing unobserved combinations improves normalized accuracy by 15.8 points over restricting the same router to observed routes. \method{} has also demonstrated training sample efficiency for new routes and components and generalization to unseen benchmarks.
% Link only in the arXiv version; it would de-anonymize the ICLR submission.
\ifarxiv Our code and data are open-sourced at \url{https://github.com/hmarkc/HM-Router}.\fi
\end{abstract}

\section{Introduction}

LLM-based agents are becoming increasingly capable and are being adopted across a broad range of complex tasks~\citep{yao2022react, merrill2026terminal, xie2024osworld, sun2026agentslastexam}.
Deploying LLMs involves choosing among models with different capabilities and costs, motivating routing methods that select a suitable model for each query~\citep{ong2025routellmlearningroutellms}.
In agentic settings, however, the model is only one component of the deployed system. Another design axis is the \emph{harness}: the surrounding scaffold that governs how a model interacts with tools, observations, and the environment~\citep{yang2024swe}.
Manually designed and automatically optimized harnesses create many ways to deploy the same model~\citep{yang2024swe, zhang2025aflow, lee2026meta}, expanding the routing decision to include both components.
This motivates \emph{model--harness co-routing}: jointly selecting a model and harness for each query based on compatibility, performance, and cost.

\begin{figure}[t]
\centering
\includegraphics[width=\linewidth]{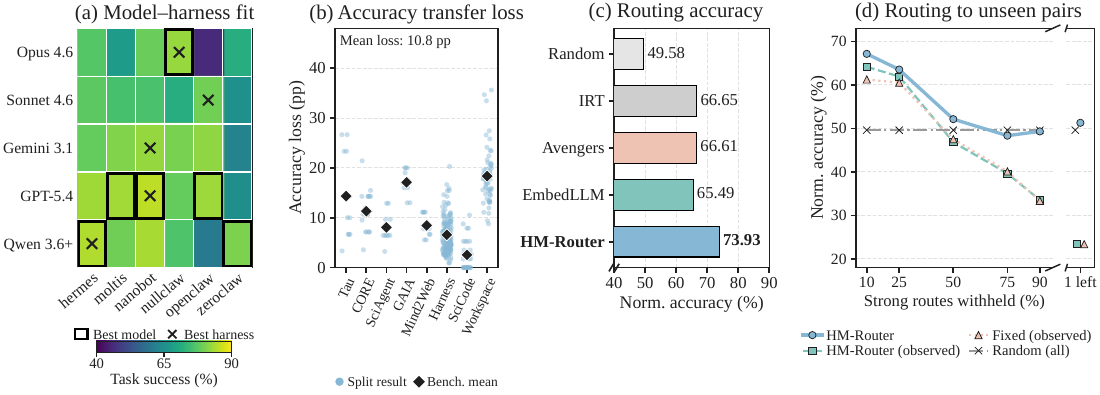}
\caption{\textbf{Joint routing: motivation and benefits.} (a) HarnessBench accuracy: boxes mark the best model per harness, crosses the best harness per model. (b) Accuracy loss when transferring oracle model choices across harnesses; dots show split results, diamonds benchmark means. (c) Mean routing accuracy across 12 benchmarks (Table~\ref{tab:routing}). (d) Routing with strong routes' training histories withheld (Table~\ref{tab:unseen}).}
\label{fig:intro}
\end{figure}

Model--harness co-routing introduces two challenges.
\textbf{Challenge 1: Model--harness compatibility.} Existing routers primarily select models~\citep{zhuang2025embedllm, song2025irt, feng2025graphrouter}, but a model's performance also depends on the harness with which it is paired. The highest-scoring model changes across harnesses (Figure~\ref{fig:intro}a). Even when we select the best model for each query under one harness, those choices can become suboptimal under another, showing that model selection must account for the harness (Figure~\ref{fig:intro}b). Joint routing must therefore capture model--harness compatibility to identify suitable pairs for each query.
\textbf{Challenge 2: Generalization to unobserved combinations.} The number of candidate routes grows as the product of the model and harness counts, making execution coverage increasingly difficult as either set expands. Treating each pair as an independent routing option ties learning to that pair's own execution history, even when its components have been evaluated independently in other combinations. This creates a coverage bottleneck: potentially strong routes may be missing from direct evidence, and excluding them can sacrifice routing accuracy (Figure~\ref{fig:intro}d).
The challenge is to transfer evidence across shared components to assess and select unobserved combinations.
Together, these challenges raise the question:

\emph{How can we effectively route queries across the joint space of models and harnesses?}

To study this question, we curate \bench{} from 12 existing public agent benchmarks, comprising 293 routes, 73 models, and 25 harnesses (Appendix~\ref{sec:benchmark}).
Using this testbed, we introduce \method{}, a routing method that learns separate representations for models and harnesses and composes them to score each candidate route. Shared component factors allow observations of one route to inform other routes containing the same model or harness. An explicit interaction term, inspired by canonical polyadic (CP) tensor decomposition~\citep{KoBa09}, allows harness preferences to depend on both the model and the query.

\method{} exceeds the strongest evaluated learned router by 7.3 percentage points in mean accuracy and leads the cost comparison at every evaluated budget. Withholding route outcomes clarifies the practical value of shared representations: the scorer can retain unobserved combinations as candidates, increasingly preserving performance as strong observed routes are removed.

Our contributions are threefold:
\begin{enumerate}
    \item We formulate query-level routing over the joint model--harness space and curate \bench{}, a controlled evaluation suite assembled from 12 existing public agent benchmarks, including a six-benchmark subset for cost analysis.
    \item We introduce a routing method that learns separate model and harness representations to share information across routes, with an explicit interaction term capturing how their compatibility varies with the query.
    \item We demonstrate that \method{} supports both effective route selection and performance prediction, using shared component evidence to extend these capabilities to model--harness combinations with limited or no execution history.
\end{enumerate}

\section{Background and Related Work}

\paragraph{Agent harnesses and their optimization.}
LLM agents interleave reasoning, tool use, and interaction with external environments~\citep{yao2022react}. A \emph{harness} orchestrates this process by managing prompts, observations, memory, and execution. Coding harnesses include Claude Code~\citep{anthropic2025claudecode}, Codex CLI~\citep{openai2025codexcli}, and Gemini CLI~\citep{google2025geminicli}. Research examples include SWE-agent's interface for navigating, editing, and testing code~\citep{yang2024swe} and Voyager's curriculum, skill library, and feedback loop for refining executable programs~\citep{wang2023voyager}. Recent methods automate harness evolution through approaches such as synthesizing code from environment feedback with AutoHarness~\citep{lou2026autoharness}, revising harness code from execution feedback with Meta-Harness~\citep{lee2026meta}, and guiding component edits with structured execution evidence in Agentic Harness Engineering~\citep{lin2026agentic}. These methods expand the space of harnesses, motivating joint selection of a model and harness for each query.

\paragraph{Model routing.}
Model routing balances quality and cost through query-level model selection, as in RouteLLM~\citep{ong2025routellmlearningroutellms}, or cascades, as in FrugalGPT~\citep{chen2023frugalgpt}; RouterBench provides a common evaluation framework~\citep{hu2024routerbench}. Most closely related, EmbedLLM learns compact model representations for routing and performance prediction~\citep{zhuang2025embedllm}, while IRT-Router models latent model abilities and query difficulty~\citep{song2025irt}. GraphRouter captures relationships among tasks, queries, and models, and supports newly introduced models~\citep{feng2025graphrouter}. These approaches learn variation across models without considering harnesses.

\paragraph{Agentic routing.}
Recent work incorporates execution evidence into model selection: Agent-as-a-Router accumulates verified execution experience for coding tasks~\citep{zhou2026agent}, SWE-Router uses partial trajectories to decide whether to escalate to a stronger model~\citep{son2026swe}, and OpenSquilla selects individual models or complementary ensembles conditioned on the current harness state~\citep{liu2026agentic}. Along the harness axis, Adaptive Auto-Harness evolves a tree of harnesses and selects a branch for each incoming task while keeping the solver model fixed~\citep{liu2026adaptive}. These works select models or adapt harnesses; \method{} jointly selects the model and harness for each query, learning their compatibility and reusing component evidence to score both observed and unobserved combinations.

\section{Problem Formulation}
\label{sec:setting}
\label{sec:method_problem}

Let $\mathcal{M}$ be a set of models and $\mathcal{H}$ a set of harnesses. A route $r=(m,h)$ pairs a model with the harness in which it runs. The candidate pool $\mathcal{R}\subseteq\mathcal{M}\times\mathcal{H}$ contains pairs that can be executed in the deployment setting. This pool need not include every combination: a harness may require interfaces or capabilities that some models do not support, or may integrate with only certain model providers. Feasibility is distinct from observation since a pair can be executable without having been evaluated.

The router learns from execution outcomes $y_{q,r}$ for training queries $q\in\mathcal{Q}_{\mathrm{train}}$. Available observations are indexed by $\Omega\subseteq\mathcal{Q}_{\mathrm{train}}\times\mathcal{R}$. Let $\mathcal{O}\subseteq\mathcal{R}$ contain routes with at least one observation. For a new query, the router selects
\begin{equation}
\label{eq:routing_rule}
\hat r(q)=\arg\max_{r\in\mathcal{R}}S(q,r),
\end{equation}
where $S$ estimates the suitability of a route for the query. Selection may include unobserved pairs in $\mathcal{R}\setminus\mathcal{O}$, so the candidate pool can extend beyond the available execution history.

This setting poses two challenges. \textbf{Model--harness compatibility:} a harness's suitability can vary across models, so it must be assessed together with the model it supports. \textbf{Incomplete observations:} the potential combination space grows as $|\mathcal{M}| \times |\mathcal{H}|$, while execution histories may cover only a subset. The router must therefore learn from observed pairs and use their shared components to predict unobserved combinations.

\section{\method{}: Compositional Model--Harness Routing}
\label{sec:method}

\begin{figure}[htbp]
    \centering
    \includegraphics[width=\linewidth]{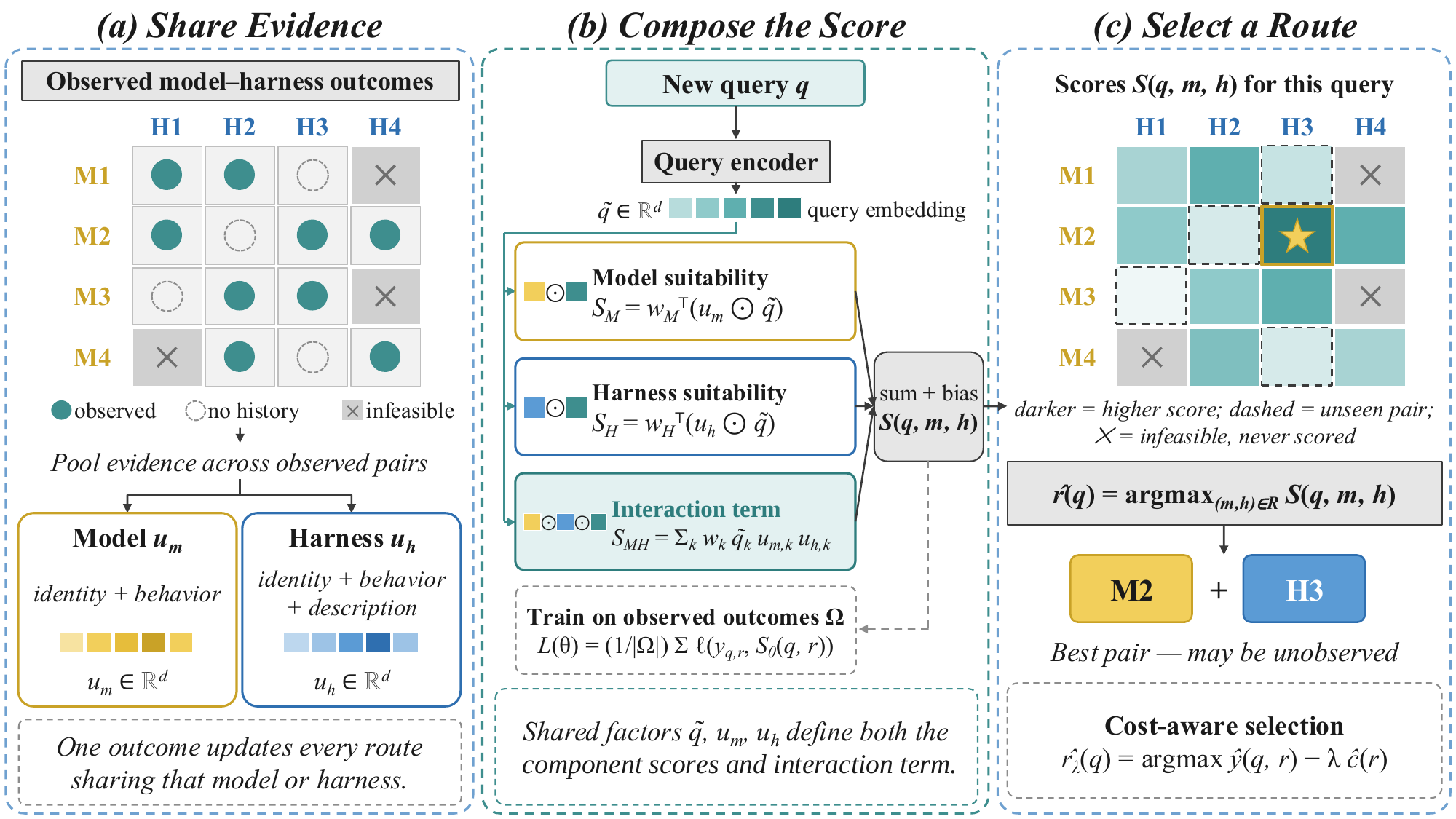}
    \caption{\method{} overview. (a) Observed outcomes inform shared model and harness representations. (b) Component and interaction scores form a query-dependent route score. (c) The router selects the highest-scoring feasible pair, including unobserved combinations, optionally accounting for execution cost.}
    \label{fig:method_overview}
\end{figure}

\method{} first shares evidence across routes through separate model and harness representations. It then models their compatibility with an interaction term built from the same factors. Figure~\ref{fig:method_overview} summarizes the resulting routing pipeline.

\subsection{Factorized route representations}
\label{sec:factorization}

The potential route space grows as $|\mathcal{M}|\times|\mathcal{H}|$, while execution histories may cover only a subset of combinations. We therefore learn separate model and harness representations, sharing each across all routes containing that component.

As illustrated in Figure~\ref{fig:method_overview}a, each component is represented using its identity and behavioral evidence pooled across observed pairings; harnesses additionally use a textual description. Specifically,
\begin{equation}
\label{eq:component_encoders}
\begin{aligned}
\qt &= W_q\mathbf{x}_q,\\
\um &= W_m[\mathbf{e}_m;P_m\mathbf{z}_m],\\
\uh &= W_h[\mathbf{e}_h;P_h\mathbf{z}_h;T_h\mathbf{t}_h],
\end{aligned}
\end{equation}
where $\mathbf{x}_q$ and $\mathbf{t}_h$ are frozen query and harness-description embeddings, $\mathbf{e}_m$ and $\mathbf{e}_h$ are learned identity embeddings, and $\mathbf{z}_m$ and $\mathbf{z}_h$ summarize each component's observed success vector over training queries.
Harness descriptions are shared across benchmarks. The learned affine maps produce $\qt,\um,\uh\in\mathbb{R}^{d}$.

These representations define how well each component matches the query. Before feature normalization, the component scores are
\begin{equation}
\label{eq:component_terms}
S_M(q,m)=\mathbf{w}_M^\top(\um\odot\qt),
\qquad
S_H(q,h)=\mathbf{w}_H^\top(\uh\odot\qt),
\end{equation}
where $\odot$ denotes elementwise multiplication. Their sum gives
\begin{equation}
\label{eq:component_score}
S_{\mathrm{comp}}(q,m,h)=S_M(q,m)+S_H(q,h).
\end{equation}
The encoders, identity embeddings, and scoring weights are learned jointly by predicting observed route outcomes (Section~\ref{sec:method_learning}). Each observation thus informs representations shared across routes, allowing an unobserved pair to be scored from evidence about its components.

\subsection{Model--harness compatibility}
\label{sec:compatibility}

The additive component score cannot express a preference between harnesses that changes with the model for the same query. To capture this dependence, we extend Eq.~\eqref{eq:component_score} with an interaction term $S_{MH}$, omitting a shared bias for clarity:
\begin{equation}
\label{eq:score_decomposition}
S(q,m,h)=S_{\mathrm{comp}}(q,m,h)+S_{MH}(q,m,h).
\end{equation}
The interaction term captures how a particular model and harness work together beyond their individual suitability. Section~\ref{sec:analysis} evaluates its contribution to routing accuracy on observed routes.

For pairs without training observations, $(m,h)\in\mathcal{R}\setminus\mathcal{O}$, we retain both the component and interaction terms. As described next, both use shared representations learned from the available observations, so scoring an unobserved combination requires no pair-specific parameters.

\subsection{Shared factors for the interaction term}
\label{sec:composition}

We construct the interaction term $S_{MH}$ from the query, model, and harness representations learned for the component score, preserving the sharing introduced in Section~\ref{sec:factorization}. Inspired by canonical polyadic (CP) tensor decomposition~\citep{KoBa09}, we extend their pairwise products to a three-way interaction term:
\begin{equation}
\label{eq:cp_interaction}
S_{MH}(q,m,h)
=\sum_{k=1}^{d}w_{k}\tilde q_k u_{m,k}u_{h,k}.
\end{equation}
Each coordinate contributes a model--harness compatibility pattern weighted by the query. Sharing factors lets the component scores and interaction term jointly shape the same representations. Before feature normalization, the resulting interaction score tensor has CP rank at most $d$; Appendix~\ref{sec:method_details} gives the implemented, normalized scorer.

\subsection{Learning and selection}
\label{sec:method_learning}

We optimize all learned parameters $\theta$ through the full route score:
\begin{equation}
\label{eq:training_loss}
\mathcal{L}(\theta)=\frac{1}{|\Omega|}\sum_{(q,r)\in\Omega}
\ell\big(y_{q,r},S_\theta(q,r)\big),
\end{equation}
where $\ell$ is the binary cross-entropy loss specified in Appendix~\ref{sec:method_details}. For a new component, probe executions supply behavioral evidence for representing its candidate routes.

At inference, we select the highest-scoring feasible route using Eq.~\eqref{eq:routing_rule}. When cost matters, we balance the predicted success probability $\hat y(q,r)=\sigma(S(q,r))$ against an estimated execution cost $\hat c(r)$:
\begin{equation}
\label{eq:cost_routing}
\hat r_\lambda(q)=\arg\max_{r\in\mathcal{R}}
\{\hat y(q,r)-\lambda\hat c(r)\},\qquad\lambda\geq0.
\end{equation}
Here $\hat c(r)$ is the mean cost of the observed route based on the training dataset, and the same route score serves routing with and without a cost penalty.

\begin{table}[t]
\centering
\caption{Routing accuracy (\%) over five splits. Values are oracle-normalized. \textbf{Bold} and \underline{underlined} entries mark the best and second-best method in each row.}
\label{tab:routing}
\normalsize
\setlength{\tabcolsep}{3pt}
\renewcommand{\arraystretch}{1.08}
\begin{tabular*}{\linewidth}{@{\extracolsep{\fill}}lrrrr>{\columncolor{ourstint}}r@{}}
\toprule
\textbf{Benchmark} & Random & IRT & Avengers & EmbedLLM & \ours \\
\midrule
TerminalBench 2.1 \citep{merrill2026terminal} & 77.29 & 82.13 & 83.32 & \second{85.71} & \best{86.89} \\
TauBench Airline \citep{yao2024tau} & 35.66 & 42.74 & 51.52 & \second{52.59} & \best{57.52} \\
CORE-Bench Hard \citep{siegel2024core} & 26.84 & 62.73 & 66.93 & \second{85.96} & \best{91.33} \\
ScienceAgentBench \citep{chen2025scienceagentbench} & 30.28 & 35.60 & \second{41.56} & 36.60 & \best{56.47} \\
GAIA \citep{mialon2024gaia} & 42.29 & 68.85 & \second{69.99} & 64.16 & \best{73.87} \\
Online Mind2Web \citep{deng2023mindweb} & 39.71 & 45.75 & 46.95 & \second{47.43} & \best{51.26} \\
SWE-bench Verified \citep{jimenez2024swe} & 61.65 & 86.27 & 86.12 & \second{88.06} & \best{88.22} \\
HarnessBench \citep{yao2026harness} & 79.71 & \second{87.83} & 82.95 & 86.44 & \best{89.33} \\
CORE-Bench 1.1 \citep{siegel2024core} & 86.60 & \best{95.00} & 88.33 & \second{93.61} & 93.33 \\
SciCode \citep{tian2024scicode} & 23.64 & \best{78.64} & 59.62 & 26.97 & \second{65.03} \\
SkillsBench \citep{li2026skillsbench} & 40.10 & 52.29 & \second{56.00} & 54.12 & \best{66.23} \\
WorkspaceBench-Lite \citep{tang2026workspace} & 51.24 & 61.94 & \second{66.01} & 64.21 & \best{67.69} \\
\midrule
\textbf{Mean} & 49.58 & \second{66.65} & 66.61 & 65.49 & \best{73.93} \\
% Mean (raw) & 43.26 & 54.96 & 55.90 & \second{57.29} & \best{61.80} \\
\bottomrule
\end{tabular*}
\end{table}

\section{Experimental Evaluation}
\label{sec:experiments}

We organize the main experiments around three research questions:
\begin{enumerate}
\item[\textbf{RQ1.}] Does \method{} achieve higher routing accuracy than existing methods?
\item[\textbf{RQ2.}] Does \method{} improve the cost--accuracy tradeoff in model--harness routing?
\item[\textbf{RQ3.}] Can our factored formulation generalize to unseen model--harness combinations?
\end{enumerate}

\subsection{Experimental setup}
\label{sec:eval_setup}

\paragraph{Benchmarks and protocol.}
We evaluate on 12 benchmarks from \bench{}, covering 293 routes, 73 models, and 25 harnesses. All methods receive the same frozen Qwen3-Embedding-8B query features and observed training outcomes, using five fixed 70/30 query splits without per-benchmark tuning. We report test accuracy divided by the split's full-pool oracle accuracy, then average over splits and benchmarks; tables and figures express this ratio as a percentage. Appendix~\ref{sec:benchmark} details benchmark coverage, route construction, and the exact model--harness pairs.

\paragraph{Baselines.}
We use Random, IrtNet (\textbf{IRT})~\citep{song2025irt}, Avengers~\citep{zhang2025avengers}, and EmbedLLM with atomic route identities~\citep{zhuang2025embedllm}. Avengers-Pro (\textbf{Av-Pro})~\citep{zhang2025beyond} is used in the cost evaluation.

\subsection{Routing performance and cost--accuracy tradeoffs}
\label{sec:main_results}

\begin{wrapfigure}{R}{0.49\linewidth}
\centering
\includegraphics[width=\linewidth]{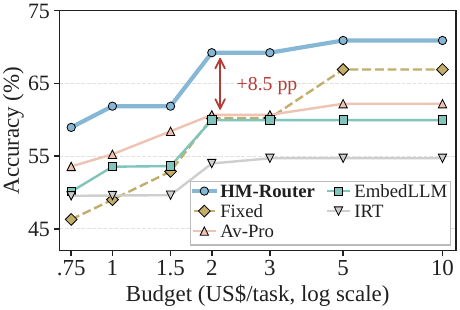}
\caption{\textbf{Cost--accuracy tradeoff.} Mean normalized accuracy by cost budget on six benchmarks.
\vspace{-15mm}}
\label{fig:cost}
\end{wrapfigure}

\paragraph{Routing quality.}
\method{} improves routing accuracy over the strongest learned baseline by 7.3 percentage points and EmbedLLM by 8.4 points (Table~\ref{tab:routing}). We compare full methods.

\paragraph{Cost--accuracy tradeoff.}
To evaluate cost, we construct \textsc{HM-Route-cost}, a subset of six benchmarks with execution-cost data. 
\method{} maintains its accuracy advantage under spending limits, as it leads at all seven evaluated dollar budgets, exceeding Av-Pro by 8.5 percentage points at \$2 per task (Figure~\ref{fig:cost}).

\Needspace{4\baselineskip}
\subsection{Routing with unobserved model--harness combinations}
\label{sec:unseen}

\paragraph{Withholding strong routes.}
Execution histories may cover only some feasible model--harness combinations. We progressively remove train-best routes' outcomes, retaining the full candidate pool at inference. Component summaries and scorer training use only the retained outcomes. For components with no remaining observations, the cold-start policy sets their identity embeddings and behavioral summaries to zero while retaining available harness descriptions. This tests a progressively degraded observed portfolio.

\begin{table}[t]
\centering
\caption{
\textbf{Routing with unobserved combinations.} Accuracy (\%) at five target removal levels and with one surviving route. \method{} (observed) selects only retained routes; Fixed (observed) selects their train-best route. Comp. Avg. (all) averages model and harness success rates over retained training outcomes, allowing hidden routes. Rankings exclude Oracle (observed).
}
\label{tab:unseen}
\setlength{\tabcolsep}{5pt}
\renewcommand{\arraystretch}{1.12}
\begin{tabular*}{\linewidth}{@{\extracolsep{\fill}}l>{\columncolor{ourstint}}rrrrr@{}}
\toprule
Removed & \ours & \shortstack{Comp. Avg.\\(all)} & \shortstack{\method{}\\(observed)} & \shortstack{Fixed\\(observed)} & \shortstack{Oracle\\(observed)} \\
\midrule
10\% & \best{67.13} & 63.52 & \second{64.15} & 61.24 & 98.30 \\
25\% & \best{63.52} & 60.22 & \second{61.92} & 60.50 & 94.22 \\
50\% & \second{52.12} & \best{52.89} & 46.82 & 47.53 & 80.29 \\
75\% & \best{48.32} & \second{40.30} & 39.62 & 40.10 & 63.48 \\
90\% & \best{49.31} & \second{37.43} & 33.54 & 33.33 & 42.22 \\
\midrule
One route left & \best{51.28} & \second{47.54} & 23.37 & 23.37 & 23.37 \\
\bottomrule
\end{tabular*}
\end{table}

\paragraph{Access to hidden routes preserves routing options.}
Allowing the same scorer to select hidden combinations improves over restricting it to observed routes: the gain is 1.6 points at 25\% removal, 15.8 points at 90\%, and 27.9 points with one observed route left (Table~\ref{tab:unseen}). At 90\% removal, \method{} also exceeds Oracle (observed). Component-based scoring can therefore recover useful options that no policy confined to the retained portfolio can select.

\paragraph{Learned scoring improves selection among unobserved combinations.}
\method{} outperforms Comp. Avg. (all) at most removal levels, with gains of 3.3 percentage points at 25\% removal and 11.9 points at 90\%. These gains show the value of learned route scores beyond component success averages, particularly when many strong routes lack execution histories. Performance is comparable at 50\% removal.

\subsection{Routing With Limited Samples}
\label{sec:cold}

\begin{figure}[t]
\centering
\includegraphics[width=\linewidth]{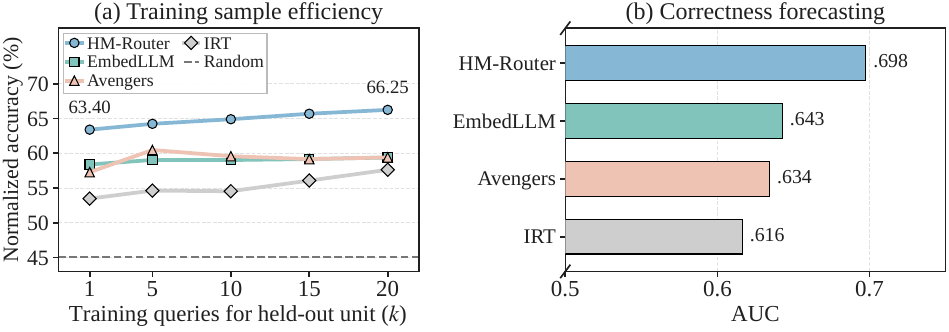}
\caption{\textbf{Sample efficiency and correctness forecasting.} (a) Oracle-normalized routing accuracy versus the number of training queries $k$ observed for a held-out route, model, or harness. (b) AUC for forecasting whether each route execution succeeds.}
\label{fig:adaptation_forecasting}
\end{figure}

\method{} achieves effective routing with small training datasets for new routes and components. To evaluate training sample efficiency, we remove all training outcomes for a route, model, or harness, then reveal its outcomes on sets of $k\in\{1,5,10,15,20\}$ training queries. Here, $k$ counts only queries observed for the held-out unit; training outcomes for other routes remain available. We aggregate results over held-out units whose removal reduces oracle accuracy by more than one percentage point.
With a training dataset of just one query for the held-out unit, \method{} exceeds EmbedLLM by 5.0 percentage points; at twenty queries, the gain increases to 6.9 points (Figure~\ref{fig:adaptation_forecasting}a). 
This sample efficiency may reflect shared model and harness representations across training samples: each outcome informs components reused across routes, allowing a small dataset for a new route or component to complement evidence from other pairings.

\subsection{Routing generalization to held-out benchmarks}
\label{sec:heldout_benchmark}

We evaluate routing on TauBench Airline, CORE-Bench Hard, and SciCode using only training samples from other benchmarks. Baselines learn a separate representation for each model--harness pair, whereas \method{} learns model and harness representations separately and can combine them to score previously unseen pairs.

\begin{figure}[t]
\centering
\includegraphics[width=\linewidth]{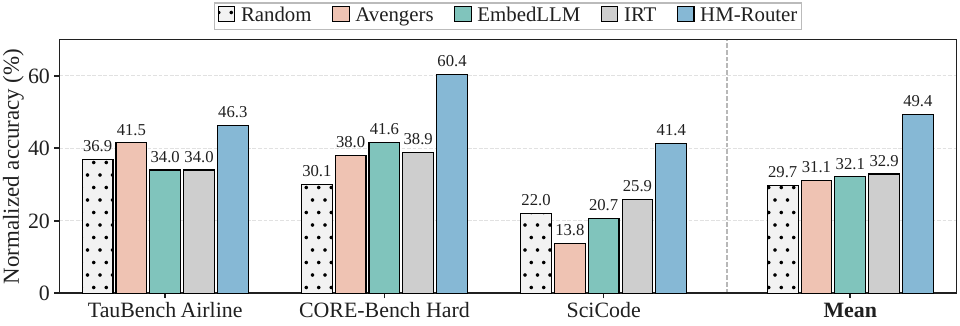}
\caption{\textbf{Routing on held-out benchmarks.} Oracle-normalized accuracy (\%) is reported. Random is the uniform routing expectation. }
\label{fig:heldout_benchmark}
\end{figure}

% TauBench Airline contains 18 routes involving 12 models and two harnesses. All components appear in the training benchmarks, but only 16 pairs do. Baselines select among these 16 pairs, while \method{} can select all 18. 
\method{} achieves the highest accuracy on all three benchmarks (Figure~\ref{fig:heldout_benchmark}), averaging 49.4\% versus 32.9\% for IRT, the strongest baseline overall. 

\subsection{Correctness forecasting}
\label{sec:forecast}

\method{} also forecasts route success more accurately, extending its usefulness beyond selecting a single route. It exceeds EmbedLLM by 0.055 AUC (Figure~\ref{fig:adaptation_forecasting}b), indicating better discrimination between successful and unsuccessful executions. 

\subsection{What do the learned representations capture?}
\label{sec:representations}

We examine whether the learned representations capture similarities between harnesses and differences in how models and harnesses work together. The first analysis uses harness vectors $\uh$ learned across benchmarks; the second examines the model--harness compatibility captured by the interaction term within each benchmark. Both use five query splits, with all evaluated routes represented in training, so the results concern generalization to new queries.

\begin{figure}[t]
\centering
\includegraphics[width=0.49\linewidth]{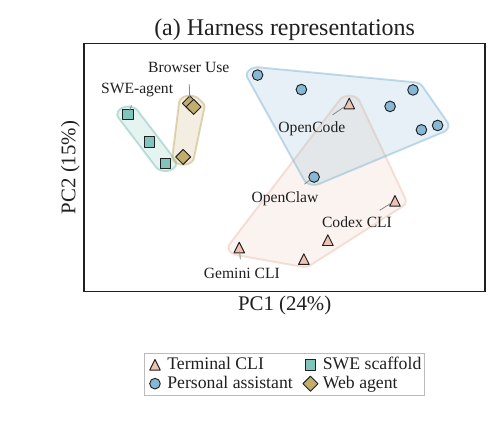}\hfill
\includegraphics[width=0.49\linewidth]{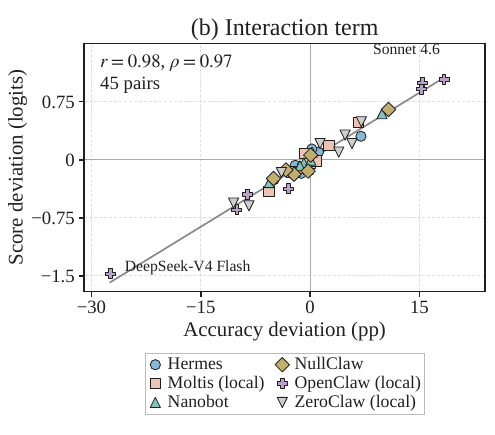}
\caption{\textbf{Harness representations and the interaction term.} (a) PCA of 18 harness vectors from four design families, centered and unit-normalized per seed, then pooled through their mean cosine matrix. Colors, markers, and shading show annotated families, unused by the router; selected harnesses are labeled. (b) Predicted score versus observed accuracy deviations for all 45 observed HarnessBench pairs, removing separate model and harness effects and averaging over test queries and five splits. The line is a descriptive linear fit.}
\label{fig:learned_representations}
\end{figure}

\paragraph{Harness representations capture similarities in design and training data.}
Harnesses with similar designs tend to have similar learned representations, even though their family labels are not used during training. Among the 18 harnesses from four families (terminal CLIs, personal assistants, SWE scaffolds, and web agents), 14 have their closest neighbor in the same family, compared with 4.3 expected when family labels are randomly shuffled (Figure~\ref{fig:learned_representations}a). This similarity may arise from both shared design and shared training samples.
Among harness pairs that share no benchmark, average cosine similarity remains higher within families than across families (0.35 versus 0.22), suggesting that shared benchmark membership alone does not explain the pattern.

\paragraph{The interaction term captures how models and harnesses work together.}
The predicted deviations identify combinations that perform better or worse than expected from their components' individual suitability. Across 45 observed HarnessBench pairs, predicted and observed deviations from these separate contributions agree closely (Figure~\ref{fig:learned_representations}b). For example, OpenClaw (local) with Sonnet 4.6 performs 18.4 percentage points above this expectation, while its pairing with DeepSeek-V4 Flash performs 27.4 points below it; the predicted deviations capture both directions. We measure these deviations by subtracting the best-fitting sum of separate model and harness effects from predicted scores and test outcomes for each query, then averaging over queries and splits. This suggests that the interaction term captures compatibility beyond individual component suitability.

\subsection{Analysis of the interaction term and routing overhead}
\label{sec:analysis}

\begin{table}[t]
\begin{minipage}[b]{0.53\linewidth}
\centering
\caption{\textbf{Interaction term ablation.} Sign agreement uses statistically resolved contrasts; Pearson measures contrast correlation.}
\label{tab:interaction}
\setlength{\tabcolsep}{3pt}
\renewcommand{\arraystretch}{1.1}
\begin{tabular*}{\linewidth}{@{\extracolsep{\fill}}lr>{\columncolor{ourstint}}r@{}}
\toprule
Metric & Component-only & \ours \\
\midrule
Accuracy (\%) & 68.51 & \best{73.93} \\
Sign agreement (\%) & 55.7 & \best{84.8} \\
Pearson & 0.20 & \best{0.49} \\
\bottomrule
\end{tabular*}
\par\vspace{0pt}
\end{minipage}\hfill
\begin{minipage}[b]{0.44\linewidth}
\centering
\caption{\textbf{Routing overhead.} Routing time per query versus agent execution.}
\label{tab:overhead}
\setlength{\tabcolsep}{3pt}
\renewcommand{\arraystretch}{1.1}
\begin{tabular*}{\linewidth}{@{\extracolsep{\fill}}lr@{}}
\toprule
Operation & Time \\
\midrule
Query encoding & 25.4\,ms/query \\
Candidate scoring & 0.2\,ms/query \\
\rowcolor{ourstint}
\textbf{Total routing} & \textbf{25.6\,ms/query} \\
Agent execution & $>100$\,s/task \\
\bottomrule
\end{tabular*}
\par\vspace{0pt}
\end{minipage}
\end{table}

\paragraph{The interaction term improves route selection.}
The interaction term improves routing accuracy by 5.4 percentage points over a version using only the component scores, with the other architecture details unchanged (Table~\ref{tab:interaction}).
This suggests that the interaction term helps distinguish the best candidates for each query.

\paragraph{The interaction term captures model-dependent harness preferences.}
The interaction term helps predict how a harness's relative performance changes across models. Across 711 observed combinations of two models and two harnesses, \method{} improves agreement with the observed direction of this change by 29 percentage points over the component-only version (Table~\ref{tab:interaction}). We measure the change as
\begin{equation}
\label{eq:rectangle}
I=[A(m_1,h_1)-A(m_1,h_2)]-[A(m_2,h_1)-A(m_2,h_2)],
\end{equation}
where $A$ is mean success on the same test queries or mean predicted success probability. Converting component-only scores to probabilities can also yield a nonzero $I$, but the higher agreement suggests that the interaction term better captures observed differences. This measures predictive agreement, not causal effects, without requiring a change in the preferred harness between models.

\paragraph{Routing adds little execution time.}
Routing introduces little latency relative to agent execution (Table~\ref{tab:overhead}). Scoring the candidate routes takes only 0.2\,ms on one NVIDIA RTX PRO 6000, while training takes less than four seconds per benchmark. Nearly all routing time comes from encoding the query, consistent with the low cost of combining shared model and harness representations.

\section{Conclusion}

Our results support joint model and harness selection when training histories cover only part of their combination space. We introduced \method{}, which shares model and harness representations across routes and captures their compatibility through an interaction term, and \bench{} to evaluate this setting. \method{} improves mean routing accuracy across 12 benchmarks and leads at all evaluated budgets on the six-benchmark cost subset. It also generalizes to unobserved combinations, effective routing from small training datasets for new routes and components, transfer to held-out benchmarks, and correctness forecasting. The interaction term improves route selection, while shared representations allow training samples from one pairing to inform others. These representations could guide future evaluation of informative combinations and harness improvements for different models.

% ICLR-required statements; omitted from the arXiv version.
\ifarxiv\else
\clearpage
\subsection*{AI use statement}

We used generative AI tools to assist with implementing parts of the evaluation scripts and interpreting aggregate results. 
We did not use these tools to generate synthetic datasets, formulate hypotheses or mathematical claims, or design the research methodology or experiments; other required-disclosure tasks are not applicable to this work. 
We also used generative AI to improve manuscript readability and revise figures. 
The authors reviewed all AI-assisted work, tested all AI-assisted code for correctness, and checked all AI-assisted text and figures against the underlying experimental data. 
We take responsibility for the final content of this work, including all text, claims, and artifacts produced with AI assistance.

\subsection*{Ethics statement}

Improved routing accuracy does not establish safe or fair agent behavior; deployment should retain safeguards for tool access and privacy and assess biases inherited from the underlying models and benchmarks.

\subsection*{Reproducibility statement}

We detail the method in Section~\ref{sec:method} and the evaluation protocol in Section~\ref{sec:eval_setup}. Appendix~\ref{sec:exp_details} provides exact model--harness pairs, scorer implementation, and hyperparameters.

\fi

% Use the bibliography style supplied with the official ICLR 2027 template.
\bibliographystyle{iclr2027_conference}
\bibliography{references}

\appendix
\section{Limitations}
\label{sec:limitations}

\bench{} covers a limited number of routes, and its model--harness grid is sparse.
Larger route pools and denser grids would allow a more complete evaluation of compatibility effects and of routing to unobserved combinations.
We will continue to add execution data to the benchmarks and to evaluate additional harnesses.

\section{Additional Experimental Details}
\label{sec:exp_details}

\subsection{Benchmark construction and coverage}
\label{sec:benchmark}

Table~\ref{tab:benchmark_pairs} lists the exact pairs of models and harnesses across benchmarks.
The cost subset comprises TerminalBench 2.1, TauBench Airline, CORE-Bench Hard, ScienceAgentBench, GAIA, and SciCode, whose route pools are re-derived by requiring a per-task cost on every query. The remaining six benchmarks either report no per-task cost or leave a degenerate cost pool.

% Generated by figures/build_benchmark_tables.py from canonical evaluation pools.
\begingroup
\setlength{\tabcolsep}{4pt}
\renewcommand{\arraystretch}{1.08}
\begin{longtable}{@{}>{\raggedright\arraybackslash}p{0.22\linewidth}>{\raggedright\arraybackslash}p{0.24\linewidth}>{\raggedright\arraybackslash}p{\dimexpr0.54\linewidth-4\tabcolsep\relax}@{}}
\caption{\textbf{Exact model--harness pairs in the retained pools.}}\label{tab:benchmark_pairs}\\
\toprule
Benchmark & Harness(es) & Model identifiers \\
\midrule
\endfirsthead
\multicolumn{3}{l}{\textbf{Table \thetable{}: Exact model--harness pairs (continued)}}\\
\toprule
Benchmark & Harness(es) & Model identifiers \\
\midrule
\endhead
\midrule
\multicolumn{3}{r}{\textit{Continued on next page}}\\
\endfoot
\bottomrule
\endlastfoot
TerminalBench 2.1 \citep{merrill2026terminal} & Claude Code \citep{anthropic2025claudecode} & \texttt{claude-fable-5}; \texttt{claude-opus-4-7}; \texttt{claude-opus-4-8}; \texttt{claude-sonnet-5}; \texttt{glm-5-1} \\
\addlinespace[4pt]
TerminalBench 2.1 & Codex \citep{openai2025codexcli} & \texttt{gpt-5-5}; \texttt{gpt-5-6-luna}; \texttt{gpt-5-6-terra} \\
\addlinespace[4pt]
TerminalBench 2.1 & Gemini CLI \citep{google2025geminicli} & \texttt{gemini-3-1-pro-preview}; \texttt{gemini-3-pro-preview} \\
\addlinespace[4pt]
TerminalBench 2.1 & Terminus 2 \citep{merrill2026terminal} & \texttt{claude-fable-5}; \texttt{claude-opus-4-7}; \texttt{gemini-3-1-pro-preview}; \texttt{gemini-3-pro-preview}; \texttt{gpt-5-5} \\
\addlinespace[4pt]
TauBench Airline \citep{yao2024tau} & HAL Generalist \citep{kapoor2026holistic} & \texttt{claude-opus-4}; \texttt{claude-opus-4-1}; \texttt{claude-sonnet-3-7}; \texttt{deepseek-r1}; \texttt{deepseek-v3}; \texttt{gemini-2-0-flash}; \texttt{gpt-4-1}; \texttt{gpt-5}; \texttt{o4-mini} \\
\addlinespace[4pt]
TauBench Airline & Taubench ToolCalling \citep{yao2024tau} & \texttt{claude-opus-4-1}; \texttt{claude-sonnet-3-5}; \texttt{claude-sonnet-3-7}; \texttt{gemini-2-0-flash}; \texttt{gpt-4-1}; \texttt{gpt-4o}; \texttt{gpt-5}; \texttt{o3}; \texttt{o4-mini} \\
\addlinespace[4pt]
CORE-Bench Hard \citep{siegel2024core} & CORE-Agent \citep{siegel2024core} & \texttt{claude-haiku-4-5}; \texttt{claude-opus-4-1}; \texttt{claude-opus-4-5}; \texttt{claude-sonnet-3-7}; \texttt{claude-sonnet-4}; \texttt{claude-sonnet-4-5}; \texttt{deepseek-chat-v3-1}; \texttt{deepseek-r1}; \texttt{deepseek-v3}; \texttt{gemini-2-0-flash}; \texttt{gemini-2-5-pro-preview-03-25}; \texttt{gemini-3-pro-preview}; \texttt{gpt-4-1}; \texttt{gpt-5}; \texttt{gpt-oss-120b}; \texttt{o3}; \texttt{o3-mini}; \texttt{o4-mini} \\
\addlinespace[4pt]
CORE-Bench Hard & Claude Code \citep{anthropic2025claudecode} & \texttt{claude-opus-4-1}; \texttt{claude-opus-4-5}; \texttt{claude-sonnet-4}; \texttt{claude-sonnet-4-5} \\
\addlinespace[4pt]
CORE-Bench Hard & HAL Generalist Agent \citep{kapoor2026holistic} & \texttt{claude-3-7-sonnet:thinking}; \texttt{claude-opus-4-1}; \texttt{claude-opus-4-5}; \texttt{claude-sonnet-3-7}; \texttt{claude-sonnet-4-5}; \texttt{deepseek-chat-v3-0324}; \texttt{deepseek-r1}; \texttt{deepseek-r1-0528}; \texttt{deepseek-v3}; \texttt{gemini-2-0-flash}; \texttt{gemini-2-5-pro-preview-03-25}; \texttt{gemini-3-pro-preview}; \texttt{gpt-4-1}; \texttt{gpt-5}; \texttt{gpt-oss-120b}; \texttt{o3}; \texttt{o4-mini} \\
\addlinespace[4pt]
ScienceAgentBench \citep{chen2025scienceagentbench} & HAL Generalist Agent \citep{kapoor2026holistic} & \texttt{claude-sonnet-3-7}; \texttt{deepseek-v3}; \texttt{gpt-4-1}; \texttt{o3}; \texttt{o4-mini} \\
\addlinespace[4pt]
ScienceAgentBench & SAB Self-Debug \citep{chen2025scienceagentbench} & \texttt{claude-haiku-4-5}; \texttt{claude-opus-4-1}; \texttt{claude-sonnet-3-7}; \texttt{claude-sonnet-4-5}; \texttt{deepseek-r1}; \texttt{deepseek-v3}; \texttt{gemini-2-0-flash}; \texttt{gemini-2-5-pro-preview-03-25}; \texttt{gpt-4-1}; \texttt{gpt-5}; \texttt{o3}; \texttt{o4-mini} \\
\addlinespace[4pt]
GAIA \citep{mialon2024gaia} & HAL Generalist Agent \citep{kapoor2026holistic} & \texttt{claude-haiku-4-5}; \texttt{claude-opus-4}; \texttt{claude-opus-4-1}; \texttt{claude-sonnet-3-7}; \texttt{claude-sonnet-4-5}; \texttt{deepseek-chat-v3-0324}; \texttt{deepseek-r1}; \texttt{deepseek-v3}; \texttt{gemini-2-0-flash}; \texttt{gpt-4-1}; \texttt{gpt-5}; \texttt{o3}; \texttt{o3-mini}; \texttt{o4-mini} \\
\addlinespace[4pt]
GAIA & HF Open Deep Research \citep{smolagents} & \texttt{claude-opus-4}; \texttt{claude-opus-4-1}; \texttt{claude-sonnet-3-7}; \texttt{claude-sonnet-4-5}; \texttt{deepseek-r1}; \texttt{deepseek-v3}; \texttt{gemini-2-0-flash}; \texttt{gpt-4-1}; \texttt{o3}; \texttt{o3-mini}; \texttt{o4-mini} \\
\addlinespace[4pt]
Online Mind2Web \citep{deng2023mindweb,xue2025illusion} & Browser-Use \citep{browser_use2024} & \texttt{claude-sonnet-3-7}; \texttt{claude-sonnet-4}; \texttt{deepseek-r1}; \texttt{deepseek-v3}; \texttt{gemini-2-0-flash}; \texttt{gemini-2-5-pro-preview-03-25}; \texttt{gpt-4-1}; \texttt{gpt-5}; \texttt{o3}; \texttt{o4-mini} \\
\addlinespace[4pt]
Online Mind2Web & SeeAct \citep{zheng2024seeact} & \texttt{claude-haiku-4-5}; \texttt{claude-sonnet-3-7}; \texttt{claude-sonnet-4}; \texttt{claude-sonnet-4-5}; \texttt{gemini-2-0-flash}; \texttt{gemini-2-5-pro-preview-03-25}; \texttt{gpt-4-1}; \texttt{gpt-5}; \texttt{o3}; \texttt{o4-mini} \\
\addlinespace[4pt]
SWE-bench Verified \citep{jimenez2024swe} & OpenHands \citep{wang2025openhands} & \texttt{claude-opus-4-5}; \texttt{claude-sonnet-3-5}; \texttt{claude-sonnet-4}; \texttt{devstral-small-2505}; \texttt{glm-4-5}; \texttt{glm-4-6}; \texttt{gpt-5}; \texttt{kimi-k2-0711-preview}; \texttt{qwen-3-coder-30b-a3b-instruct}; \texttt{qwen-3-coder-480b-a35b-instruct}; \texttt{skywork-swe-32b} \\
\addlinespace[4pt]
SWE-bench Verified & SWE-agent \citep{yang2024swe} & \texttt{claude-opus-3}; \texttt{claude-sonnet-3-5}; \texttt{claude-sonnet-4}; \texttt{devstral-small-2507}; \texttt{gpt-4-1106-preview}; \texttt{gpt-4o}; \texttt{kimi-k2-instruct}; \texttt{swe-agent-lm-32b} \\
\addlinespace[4pt]
SWE-bench Verified & Tools \citep{anthropic2024swebenchtools} & \texttt{claude-haiku-3-5}; \texttt{claude-opus-4}; \texttt{claude-sonnet-3-5}; \texttt{claude-sonnet-3-7}; \texttt{claude-sonnet-4} \\
\addlinespace[4pt]
HarnessBench \citep{yao2026harness} & hermes \citep{nous2026hermes}; moltis-local \citep{moltis2026}; nanobot \citep{nanobot2026}; zeroclaw-local \citep{zeroclaw2026} & \texttt{claude-opus-4-6}; \texttt{claude-sonnet-4-6}; \texttt{deepseek-v4-flash}; \texttt{gemini-3-1-pro}; \texttt{glm-5-1}; \texttt{gpt-5-4}; \texttt{kimi-k2-5}; \texttt{qwen-3-6-plus} \\
\addlinespace[4pt]
HarnessBench & nullclaw \citep{nullclaw2026} & \texttt{claude-opus-4-6}; \texttt{claude-sonnet-4-6}; \texttt{gemini-3-1-pro}; \texttt{gpt-5-4}; \texttt{kimi-k2-5}; \texttt{qwen-3-6-plus} \\
\addlinespace[4pt]
HarnessBench & openclaw-local \citep{openclaw2026} & \texttt{claude-opus-4-6}; \texttt{claude-sonnet-4-6}; \texttt{deepseek-v4-flash}; \texttt{gemini-3-1-pro}; \texttt{glm-5-1}; \texttt{gpt-5-4}; \texttt{qwen-3-6-plus} \\
\addlinespace[4pt]
CORE-Bench 1.1 \citep{siegel2024core} & CORE-Agent \citep{siegel2024core}; OpenCode \citep{sst2025opencode} & \texttt{claude-opus-4-5}; \texttt{claude-opus-4-6}; \texttt{gpt-5-4} \\
\addlinespace[4pt]
CORE-Bench 1.1 & Claude Code \citep{anthropic2025claudecode} & \texttt{claude-opus-4-5}; \texttt{claude-opus-4-6} \\
\addlinespace[4pt]
CORE-Bench 1.1 & Codex \citep{openai2025codexcli} & \texttt{gpt-5}; \texttt{gpt-5-1}; \texttt{gpt-5-2}; \texttt{gpt-5-3-codex}; \texttt{gpt-5-4} \\
\addlinespace[4pt]
SciCode \citep{tian2024scicode} & HAL Generalist Agent \citep{kapoor2026holistic}; SciCode Zero Shot Agent \citep{tian2024scicode} & \texttt{claude-sonnet-3-7}; \texttt{deepseek-r1}; \texttt{deepseek-v3}; \texttt{gemini-2-0-flash}; \texttt{gpt-4-1}; \texttt{o3}; \texttt{o4-mini} \\
\addlinespace[4pt]
SciCode & SciCode Tool Calling Agent \citep{tian2024scicode} & \texttt{claude-haiku-4-5}; \texttt{claude-opus-4-1}; \texttt{claude-sonnet-3-7}; \texttt{claude-sonnet-4-5}; \texttt{deepseek-r1}; \texttt{deepseek-v3}; \texttt{gemini-2-0-flash}; \texttt{gpt-4-1}; \texttt{gpt-5}; \texttt{o3}; \texttt{o4-mini} \\
\addlinespace[4pt]
SkillsBench \citep{li2026skillsbench} & Claude Code \citep{anthropic2025claudecode} & \texttt{claude-haiku-4-5}; \texttt{claude-opus-4-5}; \texttt{claude-opus-4-6}; \texttt{claude-opus-4-7}; \texttt{claude-sonnet-4-5} \\
\addlinespace[4pt]
SkillsBench & Codex \citep{openai2025codexcli} & \texttt{gpt-5.2-codex}; \texttt{gpt-5.5} \\
\addlinespace[4pt]
SkillsBench & Gemini CLI \citep{google2025geminicli} & \texttt{gemini-3-flash}; \texttt{gemini-3.1-pro} \\
\addlinespace[4pt]
SkillsBench & OpenHands \citep{wang2025openhands} & \texttt{claude-opus-4-7}; \texttt{claude-opus-4-8}; \texttt{claude-sonnet-4-6}; \texttt{deepseek-v4-flash}; \texttt{deepseek-v4-pro}; \texttt{gemini-3.1-flash-lite}; \texttt{gemini-3.1-pro}; \texttt{gemini-3.5-flash}; \texttt{glm-5.1}; \texttt{gpt-5.4-mini}; \texttt{gpt-5.5}; \texttt{grok-4.3}; \texttt{hy3}; \texttt{kimi-k2.6}; \texttt{minimax-m2.7}; \texttt{minimax-m3} \\
\addlinespace[4pt]
WorkspaceBench-Lite \citep{tang2026workspace} & Codex \citep{openai2025codexcli} & \texttt{gemini-3.1-pro}; \texttt{glm-5.1}; \texttt{gpt-5.4}; \texttt{grok-4.3}; \texttt{minimax-m2.7}; \texttt{qwen-3.6-plus} \\
\addlinespace[4pt]
WorkspaceBench-Lite & DeepAgent \citep{langchain2025deepagents}; Hermes \citep{nous2026hermes}; OpenClaw \citep{openclaw2026} & \texttt{gemini-3.1-pro}; \texttt{glm-5.1}; \texttt{gpt-5.4}; \texttt{grok-4.3}; \texttt{kimi-2.5}; \texttt{minimax-m2.7}; \texttt{qwen-3.6-plus} \\
\addlinespace[4pt]
\end{longtable}
\endgroup

\begin{table}[t]
\centering
\caption{\textbf{Accuracy attainable within a dollar budget} (\%). Columns give absolute US dollars per task. Each entry is the best setting affordable by realized test spend on each of the six cost benchmarks, then averaged. Bold and underlining rank non-oracle methods; a dash means no setting fits on at least one benchmark.}
\label{tab:cost}
\setlength{\tabcolsep}{4pt}
\renewcommand{\arraystretch}{1.1}
\begin{tabular*}{\linewidth}{@{\extracolsep{\fill}}lrrrrrrr@{}}
\toprule
\textbf{Method} & $\leq\$0.75$ & $\leq\$1$ & $\leq\$1.50$ & $\leq\$2$ & $\leq\$3$ & $\leq\$5$ & $\leq\$10$ \\
\midrule
Random & --- & --- & --- & --- & 40.20 & 40.20 & 40.20 \\
Fixed & 46.29 & 49.01 & 52.89 & 60.23 & 60.23 & \second{66.90} & \second{66.90} \\
Av-Pro & \second{53.58} & \second{55.25} & \second{58.41} & \second{60.67} & \second{60.67} & 62.20 & 62.20 \\
IRT & 49.54 & 49.58 & 49.63 & 53.98 & 54.70 & 54.70 & 54.70 \\
EmbedLLM & 50.11 & 53.54 & 53.63 & 59.95 & 59.95 & 59.95 & 59.95 \\
\rowcolor{ourstint}
\ours & \best{58.94} & \best{61.86} & \best{61.86} & \best{69.21} & \best{69.21} & \best{70.89} & \best{70.89} \\
\midrule
Oracle & 99.95 & 100.00 & 100.00 & 100.00 & 100.00 & 100.00 & 100.00 \\
\bottomrule
\end{tabular*}
\end{table}

\subsection{Scorer implementation and optimization}
\label{sec:method_details}

Each component $j$ carries a behavioral summary $\mathbf{z}_j$ built from the training outcomes of the routes containing it. Write $s_{q,j}$ for the mean outcome over those routes on training query $q$, and $n_j$ for how many of them have any observation. Then $\mathbf{z}_j$ concatenates three parts:
\begin{itemize}
\item the sum of $(2s_{q,j}-1)\mathbf{x}_q$ over queries with an observation, divided by the total number of training queries;
\item the mean of $s_{q,j}$ over those same queries;
\item $\log(1+n_j)$.
\end{itemize}
A component with no observations gets a zero summary and a zero identity embedding. Summaries are recomputed whenever outcomes are withheld or added, and withheld outcomes enter neither the summaries nor the loss. In probe experiments the identity embeddings stay zero, the probe outcomes update the summaries, and the interaction term stays active.

The implemented scorer separately normalizes the three product features:
\begin{equation}
\label{eq:scorer}
\begin{aligned}
S(q,m,h)=b
&+\mathbf{w}_M^{\top}\operatorname{LN}_M(\um\odot\qt)\\
&+\mathbf{w}_H^{\top}\operatorname{LN}_H(\uh\odot\qt)\\
&+\mathbf{w}_{MH}^{\top}\operatorname{LN}_{MH}(\um\odot\uh\odot\qt).
\end{aligned}
\end{equation}
Unlike the unnormalized interaction in Section~\ref{sec:composition}, this feature-dependent normalization does not guarantee CP rank at most $d$.

We fit one scorer per benchmark and split on the binary cross-entropy of $\hat y(q,r)=\sigma(S(q,r))$, using $d=32$, identity and side-feature width 16, and 250 Adam epochs with learning rate $10^{-3}$, weight decay $10^{-4}$, and Gaussian query noise $\sigma=0.15$.

\end{document}